\pdfoutput=1

\documentclass[11pt]{article}

\usepackage[]{acl2023}
\usepackage{times}
\usepackage{latexsym}
\usepackage{tabularx} 
\usepackage{longtable}
 \usepackage{amsmath} 
\usepackage{soul}
\usepackage{bbm}

\usepackage[T1]{fontenc}

\usepackage[utf8]{inputenc}
\usepackage{soul}

\usepackage{microtype}
\usepackage{amssymb}
\usepackage{graphicx}
\usepackage{makecell} 
\usepackage{booktabs} 

\title{MEANT: Multimodal Encoder for Antecedent Information}

\author{Benjamin Iyoya Irving\thanks{\texttt{irving.b@northeastern.edu}\\} \\
  Northeastern University\\
  Boston, MA, USA\\\And
  Annika Marie Schoene \\
  Northeastern University\\
  Boston, MA, USA}

\begin{document}
\maketitle
\begin{abstract}
The stock market provides a rich well of information that can be split across modalities, making it an ideal candidate for multimodal evaluation. Multimodal data plays an increasingly important role in the development of machine learning and has shown to positively impact performance. But information can do more than exist across modes--- it can exist across time. How should we attend to temporal data that consists of multiple information types? This work introduces (i) the
{\fontfamily{cmss}\selectfont
MEANT} model, a Multimodal Encoder for Antecedent information and (ii) a new dataset called \textit{TempStock}, which consists of price, Tweets, and graphical data with over a million Tweets from all of the companies in the S\&P 500 Index. We find that MEANT improves performance on existing baselines by over 15\%, and that the textual information affects performance far more than the visual information on our time-dependent task from our ablation study.

\end{abstract}

\section{Introduction}
Recently, multimodal models have garnered serious momentum, with the release of large pretrained architectures such as Microsoft's Kosmos-1 \cite{huang2023language} and OpenAI's GPT-4 \cite{openai2023gpt4}. Their general use has exploded in many  domains, such as language and image processing \cite{vilbert, Kim2021ViLTVT, huang2023language}. Particularly interesting to this study is the deployment of multimodal models on time-dependent environments, where recent successes have shown that event driven models processing multiple modalities are far more performant on stock market tasks than previously state of the art (SOTA) algorithms focusing purely on price information \cite{multimodal_stock, ZHANG2022117239}. Language data from news and social media sources have shown to greatly increase the performance of models for price prediction \cite{multimodal_stock, ZHANG2022117239, news, newcats, stocknet}. However, these approaches typically lack attention components specifically designed to process inputs with sequential, time-dependent information \cite{multimodal_stock, LSTMs, ZHANG2022117239, stocknet}. This type of data is particularly important when making predictions about stock prices or market movements, as price prediction is a time series task \cite{ZHANG2022117239, stocknet}. 

In this work, we introduce MEANT, a multimodal model architecture with a novel, temporally focused self-attention mechanism. We extract image features using the TimeSFormer architecture \cite{timesformer} to find relationships in longer range information (i.e a graph of stock prices over a month), while extracting language features from social media information to pick up more immediate trends (e.g.: Tweets pertaining to stock prices over a five day period). Furthermore, we release \textit{TempStock}, a multimodal stock-market dataset that is designed to be sequentially processed in chunks of varying lag periods.

\section{Related Work}

\paragraph{Multimodal Models for Financial Twitter Data} Several studies have employed natural language processing (NLP) techniques to financial markets, giving birth to the field of natural language-based financial forecasting (NLFF). Many of these studies have focused on public news \cite{news2, news}. However, social media presents more time-sensitive information from active investors. Thus, for short term analysis, many researchers have begun to focus on Tweets for feature extraction \cite{araci2019finbert, wutweet}, through which some have combined NLP techniques with traditional analysis on price data \cite{ml_stock_fundamental} . Since Tweets often correspond to events as they happen in real time, such data is better suited for smaller windows \cite{stocknet, ZHANG2022117239}. When working with stock market data, combining the features extracted through Natural Language Processing (NLP) methods with price data has shown promising results \cite{multimodal_stock, ZHANG2022117239, stocknet}. However, it is ineffective to feed the concatenated information to the model without encoding temporal dependencies \cite{multimodal_stock}. 

Modeling media-aware stock movements is essentially a binary classification problem. Many traditional machine learning methods have been deployed to solve it, including SVMs and Bayesian classifiers \cite{svms, bayesian, bayes_2}. More recently, researchers have applied deep learning to the problem. \citet{cnns} used a convolutional neural network to explore the impact of Tweets on the stock market. \citet{LSTMs} and \citet{selvin2017stock} then employed a recurrent architecture, specifically an LSTM, to extract relevant sentiments from Twitter data for stock market analysis, making their model multimodal, as it could handel Tweets and price information. \citet{multimodal_stock} built atop this architecture,  employing different tensor representations for their LSTM input to create more meaningful relationships between the price and Tweets data. 

\citet{stocknet} introduced StockNet, a large generative architecture built atop generative architectures, particularly the Variational Auto Encoder (VAE). StockNet represented the first deep generative model for stock market prediction \cite{stocknet}. TEANet, the most relevant work to our own, similarly used an LSTM to process their final output, but used a BERT-style transformer to extract relevant features from the Tweets \cite{ZHANG2022117239}. TEANet is a language model equipped to handle lag periods similarly to MEANT. They concatenate their language features to price data as an input for an LSTM and a subsequent softmax temporal encoding. We abandon recurrence altogether, developing a novel temporal mechanism, entirely based upon traditional self-attention methods \cite{vaswani}. The temporal processing in TEANet consists of concatenation methods similar to our own, but they do not employ attention over time. Furthermore, their model was built to handle Tweets and price inputs alone. MEANT can handle images as well, employing a dual encoder architecture similar to that of \citet{m3i}. 

\paragraph{Financial Twitter Datasets} Previous financial datasets have shown the power of Twitter data for financial analysis \cite{pei-etal-2022-tweetfinsent, araci2019finbert, multimodal_stock}. Twitter is powerful in its ability to generate real time information about the market before traditional newswires \cite{pei-etal-2022-tweetfinsent}. \citet{retail} focused on Twitter as a resource for examine financial dynamics in the retail sector. \citet{pei-etal-2022-tweetfinsent} introduced TweetsFinSent, a large corpus specifically for sentiment analysis. \citet{LSTMs} introduced a dataset consisting of Tweets and prices, where the Tweets information served as a sentiment analysis accompaniment for the price data. 
\citet{stocknet} introduced the StockNet-dataset, consisting of Tweets and price information for a selection of 88 companies over a two year period from 01/01/2014 to 01/01/2016. \citet{s&ptwitter} matched Tweets with price information from companies in the S\&P 500 dataset, which is the most similar to the TempStock dataset that we introduce below.

\section{TempStock Dataset}
\label{sec:data}
We collected a new dataset containing 1,755,998 Tweets and price information from all of the companies in the S\&P 500 from 4/10/2022 to 4/10/2023. 

From the price information, we calculated the Moving Average Convergence-Divergence (MACD) \cite{appel2005technical} for each company over a year. The MACD is built on the back of Exponential Moving Average (EMA) \cite{Brown1964SmoothingFA}. The EMA is defined as follows: 
$$
     EMA_t = (1 - \alpha) \cdot EMA_{t-1} + \alpha \cdot y_t
$$
where t represents the day of EMA and $y_t$ represents the closing price on that day, or in the case of the signal line, the MACD value on that day. $\alpha$ represents the degree of decrease, where $\alpha = \frac{2}{t + 1}$. The MACD consists of (i) an MACD line, which is the difference between the fast EMA and the slow EMA (commonly set to 12 days and 26 days respectively), (ii) a signal line, which is the EMA of the MACD line itself (usually over a 9 day period), and (iii) a histogram, which is the difference between the MACD and the signal line. The MACD indicator was chosen \footnote{For more on this, see \ref{sec:discussion}} because it has been shown to perform well against other indicators in terms of making accurate assertions about price directions \cite{appel2005technical, pat}. From our MACD data, we created graphs of the MACD indicator and the corresponding signal line over 26 day periods, which served as our image inputs to the MEANT model. A example of the graph inputs can be seen in Figure 1. We use graph inputs rather then numeric representations of the data to explore multi modality, and to take advantage of our attention based architecture (see \ref{sec:numeric}).

\begin{figure}
  \centering
  \includegraphics[width=\columnwidth]{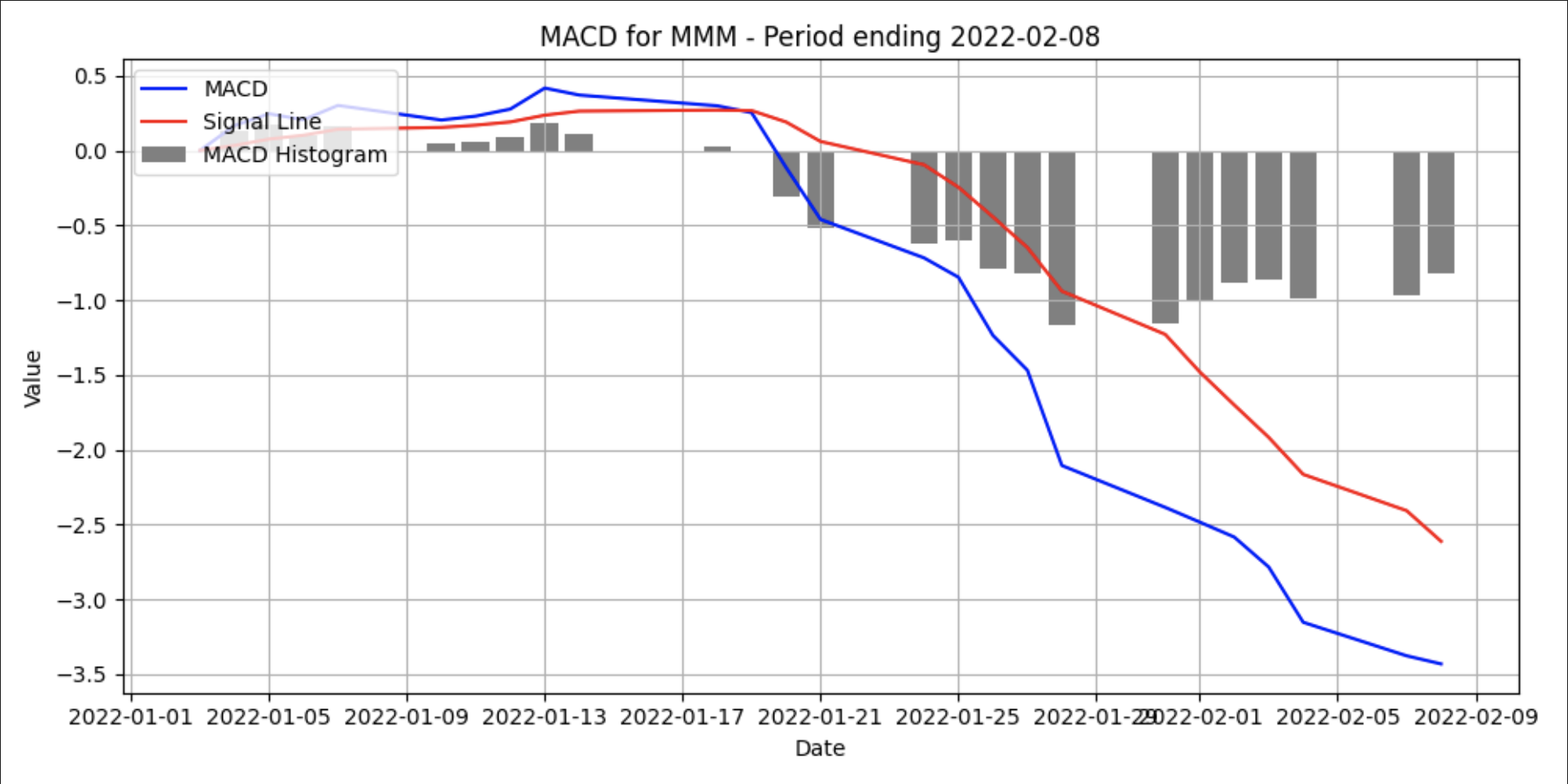}
  \caption{An example of a graph from our MACD data, which displays the MACD (in blue) and the signal line (in red) for MMM (3M) over a 26 day period. Along the x-axis, we see 11 of the dates listed, and the the y-axis shows the value of the aforementioned indicators. In each bar lies the value of the MACD histogram, which is the difference between the MACD (blue) and the Signal line (red).}
  \label{figure3}
\end{figure}

The MACD of each ticker in the subset was taken over a year period, along with the Tweets mentioning that company for each day in that period. The MACD information was gathered using the YahooFinance API \cite{yahoofinance}, and the Tweets were scraped using the snscraper \cite{snscrape} in April 2023.

\begin{table}
\centering
\begin{tabular}{lc}
\toprule
\textbf{\textbf{Description}} & \textbf{Count}\\
\hline
Total Tweets       & 1,755,998 \\
Total MACD Values  & 122,959 \\
\bottomrule
\end{tabular}
\caption{TempStock-large Raw Numbers}
\label{tab:TempStock_large_raw}
\end{table}

\paragraph{TempStock} contains Tweets, graphs, and MACD. Each input is arranged into five day lag periods leading up to target day $t$, consisting of five MACD vectors, 
$$
M=[M_{t-5}, M_{t-4}, M_{t-3}, M_{t-2}, M_{t-1}]
$$ five days of Tweets, 
$$
X=[X_{t-5}, X_{t-4}, X_{t-3}, X_{t-2}, X_{t-1}]
$$ and five images containing graphs of the MACD indicator over 26 days. $$G=[G_{t-5}, G_{t-4}, G_{t-3}, G_{t-2}, G_{t-1}]$$ For the Tweets stored daily, there were a variable amount for each ticker. We concatenated all available Tweets with [SEP] tokens in between each Tweet. These concatenations were then stored for each day in the lag period, which produced great informational variation across tickers and across days. Each MACD vector $M_i$ contains the $EMA_{12}$,  $EMA_{26}$, Signal line $s_i$, MACD histogram $h_i$, and MACD value $m_i$ for that day.
$$
M_{t-i=5,…,1} = [EMA_{12}^i,  EMA_{26}^i, s_i, h_i, m_i]
$$
In order to separate the dataset into positive and negative signals, we chose to use the MACD signal cross strategy \cite{appel2005technical}. Data points were classified as \textit{positive} if the MACD value on our preceding day to target day $t$, $m_{t-1}$, was below the Signal $s_{t-1}$, and if the MACD on our target day $m_t$ was above our Signal $s_t$.
$$
m_{t-1} < m_{t-1} \land m_t > s_{t} 
$$

Adversely, data points were classified as \textit{negative} if $m_{t-1}$ was above the Signal $s_{t-1}$, and if $m_t$ was below $s_t$ 

$$
m_{t-1} > s_{t-1} \land m_t < s_{t} 
$$

More broadly, these crossovers represent trend reversals. A positive classification (or a buy signal) in our dataset indicates that the stock is likely to begin a positive, or \textit{bullish}, momentum trend. A negative classification indicates that the stock is beginning a negative, or \textit{bearish}, reversal, meaning that the stock price will weaken overtime. The lag periods which did not fall in either of these cases were removed, along with the lag periods in which there was insufficient tweet information. This resulted in 92.57\% of the lag periods being thrown out, with the exclusion of 41 tickers from the S\&P500 all together.\footnote{Throwing out 92\% of datapoints does not indicate that TempStock will be useless 92\% of the time. It merely shows how rare buy and sell signals are in a normal market climate. You won't want your model to act on this strategy more that around 8\% of the time. An effective model which trades on momentum should not act every day.}

For more specifics on the tickers that were included, and to what extent they contributed to TempStock in its final form, please see \ref{sec:TempStock_data_details}. The resulting dataset was surprisingly balanced, with no augmentation or oversampling required. These stocks experience similar degrees of up-trends and downtrends in the time period according to the MACD rule employed above, illustrating their stability in a good market climate \cite{s&p}.

\begin{table}[h]
\centering
\begin{tabular}{@{}lcc@{}}
\toprule
\textbf{Category}      & \textbf{Count}  & \textbf{Proportion} \\
\midrule
Positive      & 4,221 &  51.36\% \\
Negative      & 3,997 &  48.64\% \\
\addlinespace
Total         & 8,218 \\
\bottomrule
\end{tabular}
\caption{TempStock splits}
\label{tab:TempStock_small_resampled}
\end{table}

\section{MEANT \label{sec:MEANT}}

\begin{figure*}[htbp]
  \centering
  \includegraphics[width=\textwidth]{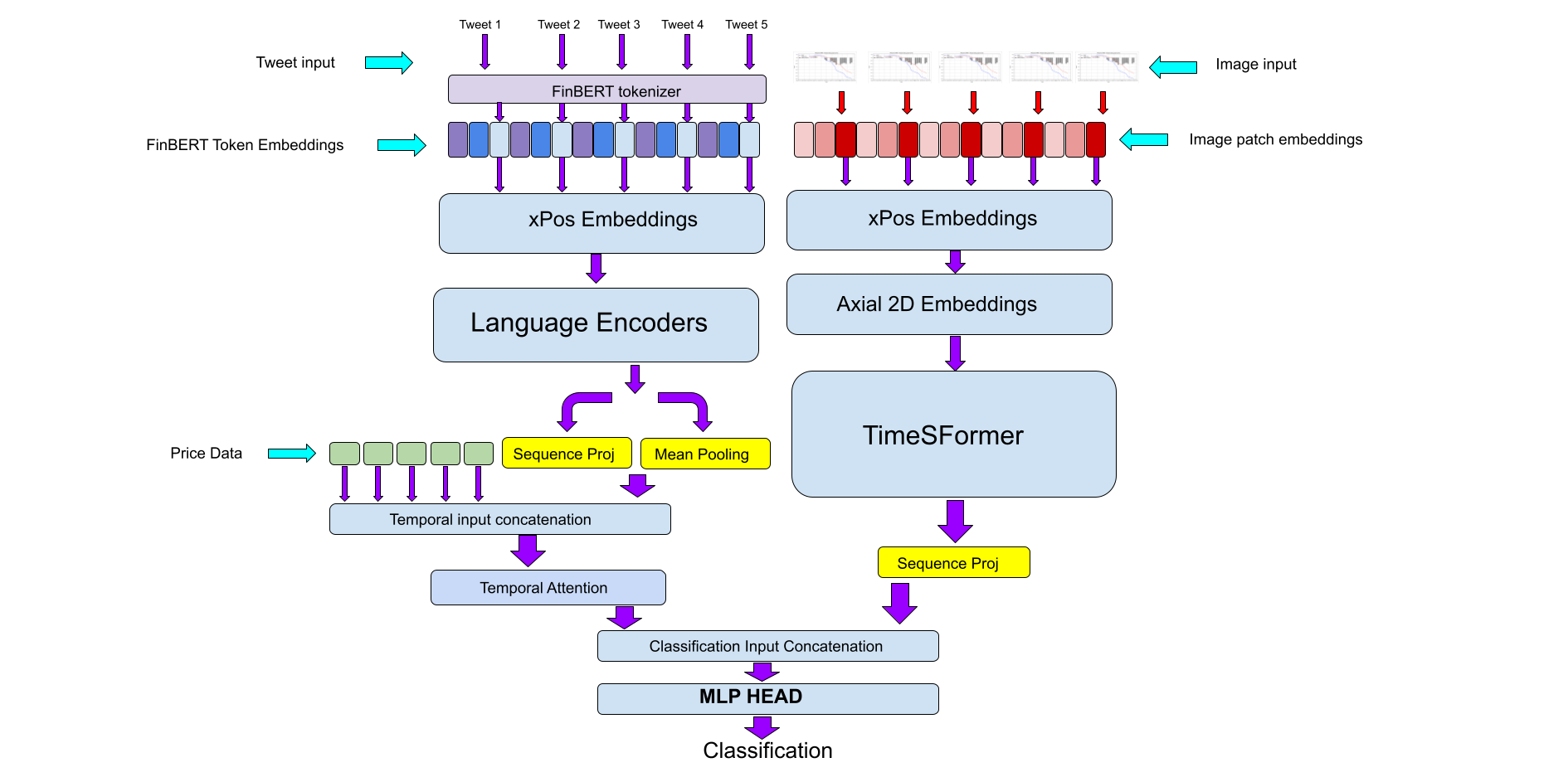}
  \caption{A schematic overview of the MEANT architecture. As seen in the diagram, the output of the language encoder is processed in two different variants: sequence projection, and mean pooling.}
  \label{fig:meant}
\end{figure*}

MEANT combines the advantages of image and language processing with temporal attention, in order to extract dependencies from multimodal, sequential information, where \ref{fig:meant} displays the full architecture. MEANT, similarly to most SOTA multimodal models \cite{multibench, Kim2021ViLTVT, vlbert, huang2023language, openai2023gpt4}, is built atop the Transformer architecture \cite{vaswani}.

\subsection{Encoder Only}
MEANT is an encoder-only model, similar to BERT \cite{bert}. Our model contains two pipelines, an image and a language pipeline. The language encoder stacks the attention mechanism with linear layers to extract relevant features from the input. Between the 2 parts of the encoder, and before the output, there is a standard residual connection, meaning that the input to that portion of the architecture is fed through added with the original input. This is done to alleviate the vanishing gradient problem \cite{pascanu2013difficulty}. The interleaved encoder structure employed by the language pipeline is inspired by the Magneto model \cite{wang2022foundation}. It makes use of sub-layer normalization, meaning that a layer norm is interleaved between the attention and linear layer components of the encoder. This architecture was chosen because it has been shown to be successful on a wide variety of uni-modal and multimodal problems \cite{huang2023language, wang2022foundation}. 

For the backbone of our image pipeline, we chose to use a variant of the TimeSFormer model \cite{timesformer}, which is an encoder model designed to handle video inputs. We chose this model because of its ability to extract dependencies in the temporal dimension. Our lag graph inputs change in place in a similar manner to a video. We altered the implementation to make use of the interleaved layernorm strategy from Magneto, and used different positional embeddings. In earlier iterations of the model, we used ViT encoders, and fed the outputs of our image pipeline to our temporal attention mechanism along with our Tweets. We found this to be less performant (see \ref{tab:TempStock_MEANT_variant}).

\subsection{Token and Patch Embeddings}
Before being fed to the attention mechanism, the two input types have to be prepared for processing using two different embedding strategies. The Tweets in MEANT are tokenized using the FinBERT tokenizer \cite{araci2019finbert} and we use the FinBERT pretrained word embedding layer.

The images are first transformed into tensors of rgb values and reshaped to a manageable size. MEANT handles input image sizes of 3 x 224 x 224, where 3 represents the number of channels and the subsequent dimensions are the height and width respectively. TimeSFormer breaks down the vectors using the patch embedding strategy from the original vision transformer \cite{Dosovitskiy} \cite{timesformer}.

\subsection{Positional Encoding}
In MEANT, the language and vision encoders use different variants of the rotary embedding \cite{rotary}. The language encoder uses the $xPos$ embeddings \cite{xPos}, while the TimeSFormer uses both rotary and axial 2-D embeddings \cite{rotary}. In axial 2-D embeddings, the angle $\theta$ of rotation is altered according to the following equation:
$$
    \theta_i = i * floor(d / 2) * pi
$$

\subsubsection{Temporal Encoder}
We developed two different variants of our temporal encoding pipeline, which work better in different cases: temporal attention with mean pooling, and temporal attention with sequence projection.

In both cases, the outputs of our language encoders $L_{out}$ are tensors of the shape $b \times l \times s \times d_l$, where $b$ denotes the batch size, $l$ denotes the lag period, $s$ is the sequence length, and $d_l$ is the dimension of each encoded language token. For temporal attention with mean pooling, we use mean pooling along the $s$ dimension:
\begin{equation}
L_{seq} = mp(L_{out}) = \frac{1}{s} \sum_{i=1}^{s} L_{out}[:, :, i, :]
\end{equation}

For temporal attention with sequence projection, we use a parameterized projection matrix to reduce $L_{out}$ along the $s$ dimension:
\begin{equation}
L_{seq} = sp(L_{out}) = \text{GELU}(\text{layNm}(W_{sl}(L_{out}^T) + b_{sl}))
\end{equation}
$W_{sl} \in \mathbb{R}^{s \times 1}$ represents our reduction weights for the language encoding. Essentially, we are extracting a latent representation for each lag day using a single layer coupled with a non-linear layer. The benefit of this is that each lag day comes to represent a token in the sequence for the attention mechanism to process.

Both of these strategies have different trade-offs, which we discuss in section \ref{seq_discussion}. Figure \ref{fig:meant} indicates where the two variations are employed to the language encoding output. 

In both cases, $L_{seq}$ has the shape $b \times l \times d_l$. To emphasize, these are the alternate formulations for the same step:

\begin{equation}
L_{seq} = 
\begin{cases} 
mp(L_{out}) & \text{(mean pooling)} \\ 
sp(L_{out}) & \text{(sequence projection)}
\end{cases}
\end{equation}

We then concatenate our $L_{seq}$ outputs to our MACD information $M$ from that five day lag period: 

\begin{equation}
    T = [ L_{seq}, M ] \in \mathbb{R}^{l \times d_t} 
\end{equation}

Where $T = [T_{t-5}, T_{t-4}, T_{t-3}, T_{t-2}, T_{t-1}]$.  $T$ has the shape $b \times l \times d_T$, where $d_T$ is the temporal dimension, which is the sum of $d_l$, and MACD price length, which is 5. $T$ signifies our inputs for the temporal encoder. In the vanilla implementation of the MEANT model, the temporal dimension is 773.  

We then pass our outputs $T$ to the temporal attention mechanism. At this point in the pipeline, relevant text features have been extracted for each trading day in relation to themselves, not to one another. The temporal attention mechanism focuses on the day before our target, or the last day in our lag period input, and its relationship to the preceding days. We put extra emphasis on this final day because its when the stock has its final movement. The inspiration for this increased focus on the final day in the lag period comes from previous work \cite{ZHANG2022117239}\cite{stocknet}, which rely on simpler, softmax-focused informational dependencies between auxiliary trading days. At inference time, we would want our model to extract a pattern from the preceding days, and act on current day with a sense of what will happen the next day, choosing to produce a buy or sell recommendation. 
MEANT does this by using a strategy we call \textit{Query-Targeting}, in which the query matrix of the attention mechanism is produced from the target day alone. To produce our query-targeted matrix $Q_t$, we first extract the pre-target day vectors from our $T$ input, $T_{t-1}$, which are of the shape $b \times 1 \times d_T$. We then multiply our learned $q$ matrix by this value to produce $Q_t$.
\begin{equation}
    Q_t = dot(T_{t-1}, q)
\end{equation}

The key and value matrices, $K$ and $V$, are calculated normally over all of $T$. The attention computation then proceeds normally with our $Q_t$, $K$, and $V$ matrices.

\begin{equation}
\small
    T_{lang} = tempAtten(Q_t, K, V) = softmax\left(\frac{Q_t K^T}{\sqrt{d}}\right)V
\end{equation}

$tempAtten(Q_t, K, V)$ results in out temporal language output $T_{lang}$, which has found the temporal dependencies between our Tweets and prices in tandem. 

TimeSFormer uses a separate strategy to extract the temporal dependencies in our image inputs, called divided space-time attention (T + S) \cite{timesformer}. The following two equations are pulled directly from 
\citet{timesformer}. (T + S) uses the patch embeddings as input, similarly to the ViT \cite{Dosovitskiy}. (T + S) first executes its temporal mechanism, where each patch attends to the patch at the same location across all of the frames. 
\begin{equation}
\small
\alpha_{(p,t)}^{(\ell, a)\text{time}} = \text{SM} \left( \frac{\mathbf{q}_{(p,t)}^{(\ell, a)}}{\sqrt{D_h}} \cdot \left[ \mathbf{k}_{(0,0)}^{(\ell, a)} \left\{ \mathbf{k}_{(p,t')}^{(\ell, a)} \right\}_{t'=1,\ldots,F} \right] \right) 
\end{equation}

In the original paper, $\ell$ denotes the encoder block, $a$ refers to the attention head, $p$ is the patch, and $t$ is the current frame. 
$\alpha_{(p,t)}^{(\ell, a)\text{time}}$ is then fed back into the spatial attention mechanisms, which executes the attention computation for each patch in relation to the other patches in its same frame, similarly to \citet{Dosovitskiy}.
\begin{equation}
\small
\alpha_{(p,t)}^{(\ell, a)\text{space}} = \text{SM} \left( \frac{\mathbf{q}_{(p,t)}^{(\ell, a)}}{\sqrt{D_h}} \cdot \left[ \mathbf{k}_{(0,0)}^{(\ell, a)} \left\{ \mathbf{k}_{(p',t)}^{(\ell, a)} \right\}_{p'=1,\ldots,N} \right] \right)
\end{equation}
$\alpha_{(p,t)}^{(\ell, a)\text{space}}$ is then fed through a feed-forward network FF and added to a residual to produce our encoded image output $I_{out}$.
\begin{equation}
   I_{out} = FF(\alpha_{(p,t)}^{(\ell, a)\text{space}}) + \alpha_{(p,t)}^{(\ell, a)\text{space}}
\end{equation}

Our output $I_{out}$ will have the shape $b \times p \times d_p$, where $p$ is the number of patches, and $d_p$ is the dimension of each patch. Similarly to how we \textit{preprocess} the outputs of our language encoder $L_{out}$ before temporal encoding, we now \textit{postprocess} our image output $I_{out}$ to extract the our temporal representation akin to the class token, using a sequence projection. 

\begin{equation}
    T_{img} = GELU(layerNorm(W_{sp}(I_{out}^T) + b_{sp}))
\end{equation}
$W_{sp}$ represents our reduction weights for the pixel encoding. We do not use mean pooling for image outputs in any variant of MEANT-base. However, we did train a ViT variant of MEANT in which we experimented with mean pooling and sequence projection for the image output. See sections \ref{tab:TempStock_MEANT_variant} and \ref{tab:TempStock_seq_proj}. 

To produce our final classification output, we concatenate our temporal representations into one vector $T_{final}$. 
\begin{equation}
    T_{final} = [T_{lang}, T_{img}]
\end{equation}

WE then pass $T_{final}$ to our MLP head to produce a classification $y$.
\begin{equation}
    y = MLP(T_{final})
\end{equation}

\section{Experiments}
\label{sec:experiments}
We ran the model at three different sizes, coined MEANT-small, MEANT-large and MEANT-XL. MEANT-small contained one encoder for language and vision, along with one temporal encoder. MEANT-large consisted of twelve language and vision encoders, and one encoder for temporal attention. twelve was selected as the number of encoders used in the original BERT model \cite{bert}. MEANT-XL had 24 encoders in our language pipeline and our TimeSFormer backbone, along with one temporal encoder. Implementation details can be seen in \ref{sec:Implementations}.

\begin{table}[h]
\centering
\small
\begin{tabular}{@{}lc@{}}
\toprule
\textbf{Model}         & \textbf{Parameter Count}\\
\midrule
MEANT-base   & 48,304,272 \\
MEANT-large   & 152,367,264 \\
MEANT-XL & 265,890,528\\
\bottomrule
\end{tabular}
\caption{MEANT Parameter Count}
\label{tab:MEANT_parameter_count}
\end{table}

\subsection{Fine-tuning on downstream tasks}
We tested the viability of the MEANT architecture on two tasks. 

\subsubsection{TempStock}
\label{TempStock_description}
TempStock is a binary classification task, identifying lag periods which resulted in momentum shifts and those that did not. To further measure MEANT's performance, we ran some similar SOTA encoder-based multimodal models on TempStock. TEANet, a key inspiration for this work, was the most similar model in original purpose, so proved the most interesting benchmark. For more details on the baselines, experiment setup, input differences and model sizes please see \ref{sec:training_settings} and \ref{tab:Model parameter Counts}.
\subsubsection{Stocknet}
The most similar dataset to TempStock was the Stocknet dataset \cite{stocknet}, which consists of Tweets and price values from a selected batch of stock tickers. Stocknet is different from TempStock as it is a unimodal dataset, containing no graphical component, and is furthermore focused on binary price change rather than momentum shift (as measured by MACD crossing in TempStock). Nonetheless, Stocknet represents one of the only datasets to our knowledge organized in lag periods and is therefore relevant as a benchmark for the MEANT model. 

Since the StockNet dataset does not have a visual input, we implemented a MEANT model without the visual capabilities called MEANT-Tweet-price. We ran MEANT-Tweet-price against TEANet \cite{ZHANG2022117239} which was originally evaluated by the authors on the StockNet dataset, as well as the StockNet model itself \cite{stocknet}. Details about the StockNet task, baselines used, training settings, input differences can be found in \ref{sec:stocknet_experiment_setup}, \ref{tab:training_settings} and \ref{tab:Model parameter Counts}.

\section{Results}

Tables \ref{tab:TempStock_large_results} and \ref{tab:stocknet_results} in sections \ref{sec:TempStock-results} and \ref{sec:stocknet-results} show the results for our experiments respectively.

\subsection{TempStockLarge Experiment results}
\label{sec:TempStock-results}
Observing \ref{tab:TempStock_large_results}, we can see that MEANT-XL outperformed all other models. MEANT-large performed comparably, coming in second for all three of those categories. The MEANT results in \ref{tab:TempStock_large_results} use sequence projection, which performed better in this task (see \ref{tab:TempStock_seq_proj}).  

Interestingly, TEANet outperformed MEANT-base. TEANet was followed closely by the LSTM baseline, which due to TEANet being built atop an LSTM backbone \cite{ZHANG2022117239}, and that the LSTM takes advantage of temporal information (the MACD values $m_{t-i}$ over all of the lag days). The MLP baseline outperforms all other BERT-based models. This illustrates the importance of the price information (further confirmed in \ref{tab:TempStock_MEANT_variant}) and attention without \textit{Query-Targeting} does not perform well. 

ViLT outperforms VL-BERT with and without the price modification. ViLT has a more similar encoding structure to MEANT, taking advantage of the patch embedding strategy, which is likely one reason for its performance advantage over VL-BERT. Since both of VL-BERT and ViLT are not designed to process lag periods, the models were at a severe disadvantage in terms of extracting temporal dependencies in the information they were given. 

For a more in depth examination of how each modality affected performance, see \ref{sec:ablation-study}.
\label{sec:TempStocklarge-results}
\begin{table}[t]
\centering
\small
\begin{tabular}{@{}lccc@{}}
\toprule
\textbf{Model}       & \textbf{F1}     & \textbf{P} & \textbf{R} \\
\midrule
FinBERT     & 0.5047 & 0.5047 & 0.5047 \\
BERT     & 0.5321 & 0.5300 & 0.5318 \\
VL-BERT     &  0.3415 & 0.2593 & 0.5000 \\
VL-BERT-price  & 0.3249 &  0.2407   & 0.5000 \\
ViLT        & 0.5483 & 0.5554   & 0.5524 \\
ViLT-price  & 0.6813 &  0.6814 & 0.6816 \\
TimeSFormer        & 0.3415 & 0.2593 & 0.5000 \\
MLP  & 0.7124&  0.7145   & 0.7122 \\
LSTM  & 0.7623 &  0.7622  & 0.7623 \\
TEANet     & 0.7898 & 0.8198    & 0.7979 \\
\addlinespace
MEANT-base  & 0.7815 & 0.7917 & 0.7812 \\
MEANT-large & 0.8351 & 0.8399 & 0.8343 \\
MEANT-XL & 0.\textbf{8440} & \textbf{0.8497} & \textbf{0.8430} \\
\bottomrule
\end{tabular}
\caption{TempStock-Large Experiment Results, using Precision (P), Recall (R), and F-1 scores.}
\label{tab:TempStock_large_results}
\end{table}

\subsection{Stocknet results}
\label{sec:stocknet-results}

\begin{table}[h]
\centering
\small
\begin{tabular}{@{}lcccc@{}}
\toprule
\textbf{Model}       & \textbf{Acc\%} & \textbf{F1}     & \textbf{P} & \textbf{R} \\
\midrule
MLP     & 50.17 & 0.49 & 0.50 & 0.50 \\
LSTM     & 54.76 & 0.47 & 0.59 & 0.54 \\
FinBERT      & 46.17   & 0.29 & 0.21  & 0.50  \\
BERTweet      & 49.20   & 0.32 & 0.24  & 0.50  \\
StockNet      & 57.53   & 0.57  & 0.58  & 0.57  \\
TEANet        & 67.75   & 0.68  & 0.67  & 0.68  \\
\addlinespace
M-Tweet-price-base &  79.92  & 0.79  & 0.80 &  0.79 \\
M-Tweet-price-large &  81.35  & 0.81 & 0.81 &  0.81 \\
M-Tweet-price-XL & \textbf{82.15} & \textbf{0.82} & \textbf{0.82} & \textbf{0.8211} \\
\bottomrule
\end{tabular}
\caption{StockNet-dataset experiment results using Precision (P), Recall (R), F-1 scores and testing accuracy (Acc).}
\label{tab:stocknet_results}
\end{table}

Looking at \ref{tab:stocknet_results}, MEANT-Tweet base and MEANT-Tweet-large, both using mean pooling, outperform all other models by a significant amount on the StockNet task. MEANT-tweet-XL outperformed TEANet, the previous SOTA on the StockNet dataset, by 15\%. We ran our own implementation of the TEANet model on the task following their descriptions from the paper, as we could not find publicly available code (see \ref{sec:Implementations}). The original accuracy score reported in the paper was 65.16\% \cite{ZHANG2022117239}. 

The importance of a temporal component for the StockNet task is clear. BERTweet, a typical encoder architecture without temporal support, performed abysmally. StockNet performed marginally better, but it is with the auxiliary temporal softmax mechanism in TEANet that the first true performance gain can be seen. In these runs, our mean pooling mechanism was more effective than the sequence projection strategy for our temporal encoding (see section \ref{seq_discussion} in our appendix).  

Clearly, the attention-based temporal mechanism in MEANT is the most performant for this problem. \textit{Query-Targeting} is able to extract meaningful relationships between the target day and the auxiliary trading days more effectively than previous mechanisms. There are likely a few reasons for this. Models that depend on multi-head selt-attention (MSA) can be thought of as a low pass filters, meaning that they generally tend to flatten loss landscapes \cite{park2022vision}. There are Tweets in the StockNet dataset that don't correlate to the buy signal, but because of the nature of the data collection, these are in the vast minority \cite{stocknet}. However, since we are also extracting trends that are dependent on the order of these Tweets in time, a succession of even a few outlier or irrelevant Tweets could be very damaging to the loss landscape of a more sensitive model. Our temporal attention mechanism is better able to handle the noise in the data. Furthermore, attention scales far better with parameter size, and our MEANT-XL model in particular dwarfs previous TEANet and StockNet in parameter size \cite{ZHANG2022117239, stocknet}. Larger parameter spaces tend to lead to a more nuanced loss landscape \cite{loss, goldilocks, park2022vision}. 

\subsection{Limitations}
\label{sec:discussion}
Here, we outline considerations, trade-offs and design decisions we have made:

\begin{itemize}
    \item \textbf{Dataset}
    To explore temporal information processing, we chose momentum buy signals in stock market data. We went with the MACD indicator because of its robustness, and correlation to strong positive returns against other indicators \cite{macd-evidence, pat}. The serious drawback in this choice is in the infrequency of buy and sell signals that occur, which leads to a less robust dataset. 

    We gathered our stock price information from companies in the S\&P 500. We chose this index because of its stability. However, as a result, we were unable to train our model on more extreme price patterns that are more common on obscure indexes \cite{s&p}. Thus, in the case of extreme market events that result in periods of steep decline or rise would likely confuse the model.
    
    \item \textbf{MEANT}
    The MEANT encoder is built atop the Kosmos-1 encoder architecture, that uses interleaved LayerNorms \cite{vu-etal-2022-layer}. The authors thought this to lead to increased numeric stability \cite{huang2023language}, which in turn helps prevent the exploding gradient problem. However, the inclusion of so many layerNorms in each encoder in our models can lead to an increase in bias, which eventually can lead to a serious overfitting problem \cite{layernormtradeoff}. We chose to go ahead with this risk, as previous architectures have shown the stability gains from the interleaved normalizations to allow for better scaling \cite{wang2022foundation, huang2023language}.
    
    MEANT was trained to identify buy signals and sell signals, instead of trying to classify price periods on a more nuanced scale. We chose this path for simplicity's sake. For practical use on financial data, we would likely need more levels of categorization.

\end{itemize}

\section{Conclusion and Future Work}
We introduced a multimodal encoder with a novel temporal component comprised entirely of self-attention. MEANT outperforms previous models on the StockNet benchmark by 15\%, and proves to be the most performant model on our own TempStock benchmark. To our knowledge, MEANT-XL is the largest model to be applied to StockNet, and is the first multimodal model to contain an attention mechanism to deal with data over a lag period of days. MEANT combines the realms of language, vision, and time to produce SOTA results. We would like to explore different early fusion methods in order to make MEANT more robust against other common multi modal benchmarks, and expand upon our \textit{Query-Targeting} strategy to emphasize relevant queries automatically, rather manually emphasizing any specific component such as the final lag day. We believe that the MEANT architecture has the potential to succeed on a wide variety of tasks. Furthermore, the image space that we trained MEANT on was limited. We would like to introduce more variation into our image inputs, to fully utilize the capabilities of that modality in our model. 

\section{Ethics Statement}
\paragraph{Bias and Data Privacy:} We acknowledge that there are biases in our study, including limiting our work to a specific time period, a small sample of securities and the general public, where we cannot verify they financial expertise in assessing markets. The data collected in this work will only be made available via Tweet IDs collected to protect X's users rights to remove, withdraw or delete their content. All datasets and Language Models are publicly available and were used under the license category that allows use for academic research.

\paragraph{Reproducibility:} We make all of our code publicly available upon publication on Github\footnote{\url{https://github.com/biirving/meant}}.

\paragraph{Use case:} We strongly advise against the use of our proposed model and dataset for financial decision making, including but not limited to automated or high frequency trading.


\bibliography{anthology,custom}
\bibliographystyle{acl_natbib}

\appendix

\section{Appendix}
\label{sec:appendix}

\subsection{Albation Study}
\label{sec:ablation-study}

To examine the importance of the image and language modalities respectively, we also created many variations of the MEANT model, to target each modality and different combinations of them. Thus, we had a model for each modality individually, and each combination of the three modalities. MEANT-vision-price and MEANT-Tweet-price, for instance, take in the inputs $x = [G, M]$ and $x=[X, M]$ respectively. All variants were similarly fine-tuned and evaluated on the TempStock task (\ref{TempStock_description}) over 15 epochs, with a training batch size of 16, a starting learning rate of 5e-5, the AdamW optimizer, and a cosine-annealing learning rate scheduler with warm restarts. 

\begin{table}[h]
\centering
\small
\begin{tabular}{@{}lccc@{}}
\toprule
\textbf{MEANT Ablation}        & \textbf{F1}      & \textbf{P} & \textbf{R} \\
\midrule
MEANT-base  & 0.7815 & 0.7917 & 0.7812 \\
MEANT-large & 0.8351 & 0.8399 & 0.8343 \\
MEANT-XL & \textbf{0.8440} & \textbf{0.8497} & \textbf{0.8430} \\
MEANT-base-pt  & 0.7712 & 0.8039 & 0.7654 \\
MEANT-large-pt & 0.8249 & 0.8272 & 0.8258 \\
MEANT-XL-pt & 0.8312 & 0.8322 & 0.8288 \\
MEANT-base-10  & 0.5731 & 0.5031 & 0.5631 \\
MEANT-large-10 & 0.6294 & 0.6227 & 0.6285 \\
MEANT-XL-10 & 0.6315 & 0.6321 & 0.6277 \\
M-Tweet-price & 0.7375 & 0.8168 & 0.7565 \\
M-Tweet-price-large      & 0.8305 & 0.8346  & 0.8327 \\
M-Tweet-price-XL      & 0.8337 & 0.8359  & 0.8348 \\
M-Tweet       & 0.3415 & 0.2593    & 0.5000 \\
M-Tweet-Large       & 0.4213 & 0.4176   & 0.5328 \\
M-Tweet-XL     & 0.5013 & 0.4776   & 0.5593 \\
M-vision-price      & 0.3249 & 0.2407  &  0.5000\\
M-vision-price-large      & 0.5237 & 0.3815  & 0.5769 \\
M-vision-price-XL & 0.7104 & 0.7103  & 0.7104 \\
M-vision-no-price & 0.3415  & 0.2593 & 0.5000 \\
M-vision-no-price-l & 0.3415  & 0.2593  & 0.5000 \\
M-vision-no-price-XL & 0.3725 & 0.3293  & 0.5784 \\
M-price-large & 0.7376 & 0.7285  & 0.7479 \\
MEANT-ViT-Large & 0.7477 & 0.7844 & 0.7639 \\
MEANT-no-lag  & 0.5942 & 0.5145 & 0.5523 \\
\bottomrule
\end{tabular}
\caption{TempStock MEANT-variant Results, using Precision (P), Recall (R), and F-1 scores.}
\label{tab:TempStock_MEANT_variant}
\end{table}
Examining \ref{tab:TempStock_MEANT_variant}, we see that MEANT-XL exhibited the best performance in F1, precision, and recall. What is perhaps more interesting about these results is examining the performance of MEANT-Tweet-price vs MEANT. The performance drop-off from MEANT-large to MEANT-Tweet-price-large is only about 0.046 in F1 score. Yet MEANT-vision-price-large exhibits a performance drop off of 0.31 from MEANT-large. These results indicate that the Twitter inputs contain features which are more indicative of momentum changes in the MACD indicator than the long-range graph inputs. There are likely many reasons for this phenomena, the primary of which being that stock prices seem to flucuate on short time periods \cite{ZHANG2022117239} \cite{stocknet}. As such, the long range information encoded in our graphs likely just introduces noise which degrades model performance.

We find that the five day lag period seems to be ideal for price prediction problems. Testing on a lag period of 10, MEANT performance drops considerably. Tweet information in particular is known to be short range, users tending to contribute information predicated upon immediate trends \cite{araci2019finbert}\cite{bertweet}\cite{stocknet}\cite{ZHANG2022117239}. As such, introducing information over a longer time period only serves to weaken the relevant signals our model is looking for. We also tested MEANT without a lag period. Single-day data proves insufficient. As such, 5 appears to be in our Goldilocks zone. 

The price modality clearly important to the performance of the MEANT model. MEANT-price, which only takes in $M$ as an input, performs admirably, vastly outperforming MEANT-Tweet-large and MEANT-vision-large, which take in $X$ and $G$ as inputs respectively.

Removing the price modality from MEANT-Tweet model reduces performance by 0.40 in F1 for the large models. For the vision models, the reduction in performance for the large models is 0.36. In fact, performance seems to collapse completely, with the base, large, and XL models of MEANT-vision achieving the same abysmal performance. The price information is what determines to the labels, so it makes sense that our model performance would be negatively affected by the removal of the $M$ inputs. 

We did try pretraining, using the TempStock raw Tweets and graphs in masked-language-modeling and masked-image-modling regimes. We found that the pretrained models performed no better on our task. The settings used for our pretraining scheme can be seen in \ref{sec:pretraining}.

In a previous iteration of the model, we used ViTs as the image backbone, and actually fed our concatenated image-tweet encoder outputs into the same temporal attention mechanism. We found that performance with this architecture was worse, likely due to the confusion in the temporal attention pass introduced by the early fusion strategy.

\begin{figure}
  \centering
  \includegraphics[width=\columnwidth]{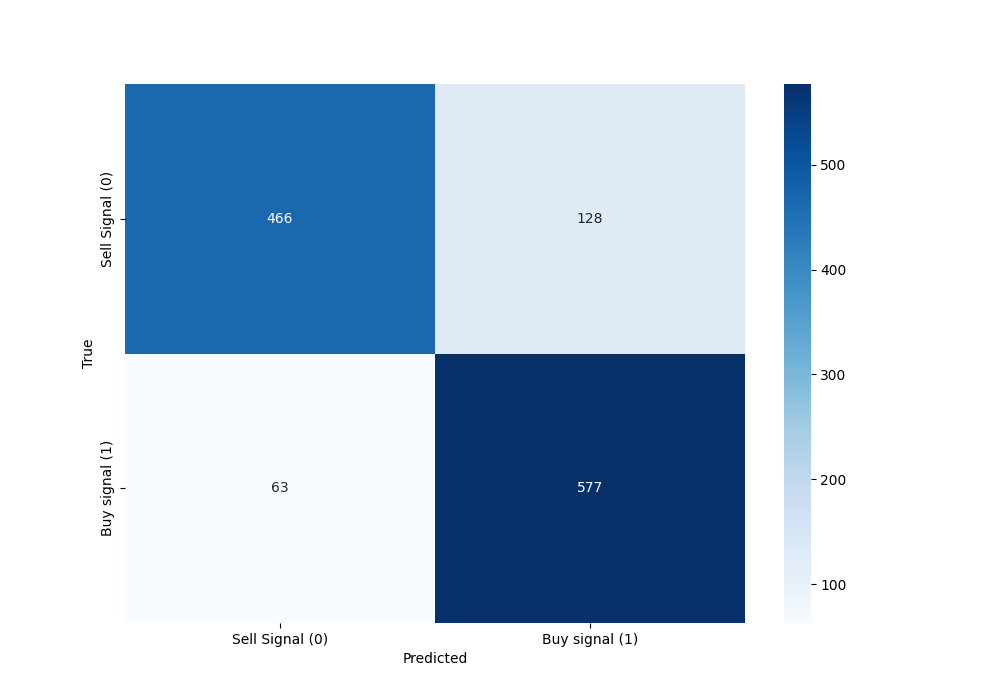}
  \caption{Confusion matrix for MEANT-XL on TempStock}
  \label{fig:MEANT_confusion}
\end{figure}

\label{sec:confusion matrices}
\begin{figure}
  \centering
  \includegraphics[width=\columnwidth]{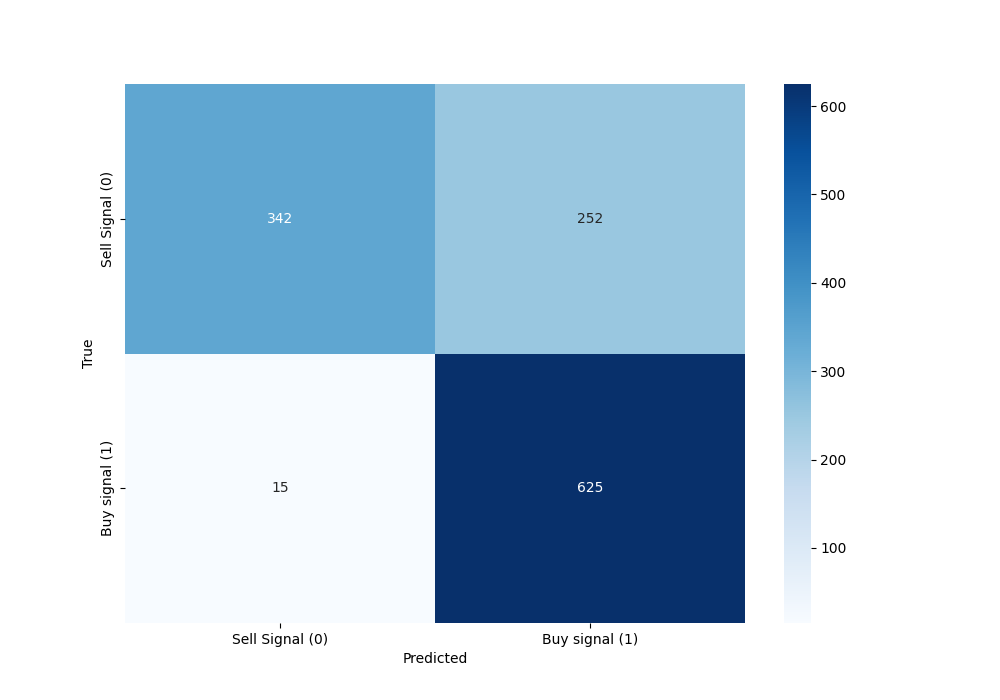}
  \caption{Confusion matrix for TEANet on TempStock}
  \label{fig:teanet_confusion}
\end{figure}

\subsubsection{Sequence Projection Vs. Mean Pooling}
\label{seq_discussion}
Using sequence projection vs mean pooling in our temporal attention mechanism had an affect on our model performance across both of our tasks. 

Looking at \ref{tab:TempStock_seq_proj}, sequence projection outperformed mean pooling for our language encoder outputs on the TempStock task by a reasonable margin, the disparity especially noticeable between MEANT-Large-MP and MEANT-Large-SP.

TempStock is built upon the MACD indicator, which relies on information over a longer time period than simple price prediction \cite{macd-evidence}, with the MACD calculation involving price averages over 12 and 26 days. Much of that information is not captured in our semantic inputs \cite{macd-evidence} which tend to correlate to short term trends of a few days or so (see \ref{tab:TempStock_MEANT_variant}). Furthermore, Tweets tend to vary widely in terms of quality \cite{araci2019finbert}\cite{stocknet}. What semantic information is pertinent to our final output must be captured with some degree of delicacy, similar to how \citet{stocknet} discerns what Tweets to throw away. A lot of the semantic input is likely just noise which confuses our model, and the parameterized extraction of important Tweets for each lag day alleviates this problem to some extent. 

\begin{table}[h]
\centering
\small
\begin{tabular}{@{}lccc@{}}
\toprule
\textbf{TempStock Seq. proj results}        & \textbf{F1}      & \textbf{P} & \textbf{R} \\
\midrule
MEANT-Large-MP    & 0.6143 & 0.6241  & 0.6173\\
MEANT-XL-MP    & 0.7983 & 0.8265   & 0.8058 \\
MEANT-Large-SP & 0.8351 & 0.8399 & 0.8343 \\
MEANT-XL-SP & \textbf{0.8440} & \textbf{0.8497} & \textbf{0.8430} \\
\bottomrule
\end{tabular}
\caption{TempStock Seq proj results, using Precision (P), Recall (R), and F-1 scores.}
\label{tab:TempStock_seq_proj}
\end{table}

Interestingly, mean pooling actually performs better than sequence projection on the StockNet task (see \ref{tab:stocknet_seq_proj}). The disparity in this case is glaring. With sequence projection, MEANT performs abysmally, essentially making random guesses with each input. There are likely a few reasons for this.

For one, the StockNet task is a binary price prediction problem, which exists on a far smaller timescale than TempStock it terms of its information. Thus, the semantic Tweet inputs are likely to contain far more robust correlations to the labels then in the TempStock problem. In other words, the Tweets have a far larger sway over StockNet performance then in TempStock (which is a phenomena observed in previous work that measures on the StockNet dataset \cite{stocknet} \cite{ZHANG2022117239}). 

Mean pooling manages to preserve spatial information, summarizing local neighborhoods (in this case, Tweets that have been encoded into different part of each sequence in $X$). A projection, on the other hand, can destroy spatial correlations in the new basis \cite{pooling_upsides}. What seems to be happening here is our learned projection is throwing away crucial Tweet information, in a problem where the Tweets have a larger importance. While the parameterization serves to intelligently extract the 'relevant' information, in the case where there is little noise in our semantic information, this parameterized projection only serves to damage performance.

\begin{table}[h]
\centering
\small
\begin{tabular}{@{}lccc@{}}
\toprule
\textbf{Stocknet Seq. proj results}        & \textbf{F1}      & \textbf{P} & \textbf{R} \\
\midrule
M-Tweet-large-MP &0.8134 & 0.8135 &  0.8133\\
M-Tweet-XL-MP & \textbf{0.8212} & \textbf{0.8225} & \textbf{0.8211} \\
MEANT-Large-SP & 0.4401 & 0.5704 & 0.5259 \\
MEANT-XL-SP  & 0.4520 & 0.5725 & 0.5303 \\
\bottomrule
\end{tabular}
\caption{Stocknet Seq proj results, using Precision (P), Recall (R), and F-1 scores.}
\label{tab:stocknet_seq_proj}
\end{table}

\subsubsection{Numeric vs Graphical Inputs}
\label{sec:numeric}
We wished to encode long range information into our model weights using images. The attention architecture in our transformer model is able to capture long range dependencies in disparate parts of images \cite{park2022vision}. Only using numeric representation would not take advantage of attention for this purpose. We have run experiments with the numeric value representations of the graphs instead of the images themselves (labeled \textit{numeric-graph} in the table below). We used an LSTM instead of TimeSformer for this input stream, and measured the performance vs MEANT.

\begin{table}[h]
\centering
\label{tab:tempstock-results}
\small
\begin{tabular}{lccc}
\hline
Model & F1 & P & R \\
\hline
MEANT-base & 0.7815 & 0.7917 & 0.7812 \\
MEANT-large & 0.8351 & 0.8399 & 0.8343 \\
MEANT-XL & \textbf{0.8440} & \textbf{0.8497} & \textbf{0.8430} \\
MEANT-base-numeric-graph & 0.7475 & 0.7382 & 0.7671 \\
MEANT-large-numeric-graph & 0.8011 & 0.7963 & 0.8078 \\
MEANT-XL-numeric-graph & 0.8119 & 0.8116 & 0.8172 \\
\hline
\end{tabular}
\caption{TempStock Graphs vs Numeric Results}
\end{table}

Keeping all other modalities the same, there is a performance drop-off of about 0.03 in F1, which seems to be reasonable empirical evidence for keeping the graph modality instead of a numeric substitute. The performance drop is likely due to the LSTM's inability to capture long range dependencies in the input.

\subsection{Pretraining}
\label{sec:pretraining}
For experimental purposes, we tried pretraining the MEANT language encoders on the TempStock Tweets.

We follow typical pretraining methods. For our language encoder, we used masked language modeling on our raw TempStock data. We trained our MEANT-small and MEANT-large language encoders on 4 NVIDIA p100 GPUs for 3 and 10 hours respectively. For MEANT-XL, we trained on an A100 GPU for 10 hours. A training batch size of 32 was used. 

For the TimeSFormer backbone, we used masked image modeling with block and channel masking. The image encoders were trained on 4 NVIDIA p100 GPUs as well, for 20 hours. We used graphs $G$ from the raw MACD data in TempStock. For these encoders, we also used a training batch size of 32.

\subsection{Training Details}
\label{sec:training_settings}
All training was done with an AdamW optimizer \cite{adamw} using betas of 0.9 and 0.999, a cosine annealing learning rate scheduler with warm restarts with 7 iterations for the first restart \cite{cosinewarm}, and an initial learning rate of $5e^{-5}$. The experiments were all run on a single NVIDIA A100 GPU. More specific settings can be seen in \ref{tab:training_settings}.
\subsubsection{TempStock Experiment Setup}
TEANet makes use of a BERT-style encoder for the Tweet inputs, but uses an LSTM on the concatenated price-Tweet data rather then relying on a pure self-attention based mechanism. Furthermore, TEANet's temporal attention is a softmax-based mechanism which uses some simple concatenation to draw relationships between the last input day and the auxiliary days. TEANet can process lag periods, but cannot process the image inputs and is thus only fed the tweet and price information $X$ and $M$. 

We also fine-tuned VL-BERT \cite{vlbert} and ViLT \cite{Kim2021ViLTVT} on TempStock. VL-BERT is an early-fusion multimodal model, that uses a Faster RCNN \cite{fastrcnn} to extract the image features, which are concatenated to the textual features before being fed to a BERT-style encoder. VL-BERT cannot process the price data, or data over the lag period, so we fed the model the graphs and Tweets from the final auxiliary day, those being $G_{t-1}$ and $X_{t-1}$ respectively.  

ViLT is a single stream encoder that uses a ViT style patch embedding on the images, concatenating these to the text embeddings before feeding the concatenated input to a BERT-style encoder \cite{Kim2021ViLTVT}. ViLT, similarly to VL-BERT, cannot process price data, or data over a lag period. So we fed the model the same inputs as VL-BERT. 

We recognized that the lack of price data could give tremendous advantages to TEANet and MEANT over ViLT and VL-BERT, as the labels of TempStock are determined directly from the price component. Thus, we added some extra functionality to our own variants of ViLT and VL-BERT models, called ViLT-price and VL-BERT-price respectively, to handle prices for better comparison of their multimodal strategies. We simply concatenated the price to our encodings of the images and Tweets before feeding the vectors into the attention mechanism. These models recieved the price, graphs, and text data from the last auxiliary day, $M_{t-1}$, $G_{t-1}$, and $X_{t-1}$ respectively.

FinBERT and BERT were simply given the Tweets $X_{t-1}$ from the final auxiliary day. For parameter comparisons, see \ref{tab:MEANT_parameter_count}

For the TempStock experiment, we used 15 epochs for all MEANT models and a train batch size of 16. We decided to run TimeSFormer on the dataset as well, giving it the images over the lag period as a vision-only baseline. For more simple baselines, we ran a simple MLP on TempStock without a lag functionality, only taking in the prices $M_{t-1}$ from the day before the target period. We also ran an LSTM \cite{LSTMs}, but with a different input of the MACD values $m_{t-i}$ for $i=1,...5$, to see if the recurrent properties could extract a pattern.

We used the lag periods from 4/10/2022-12/10/2023 for our training set, the periods from 11/10/2023-2/25/2023 as our validation set, and the periods from 2/25/2023-4/10/2023 as our test set. 

\subsubsection{StockNet Experiment Setup}
\label{sec:stocknet_experiment_setup}

The StockNet model was the predecessor to TEANet. StockNet took advantage of a similar Temporal attention mechanism, but used gated recurrent units rather then a BERT-style encoder to process the Tweets, and employed a the use of a latent representation with a variational lower bound for optimization \cite{stocknet}.

We ran BERTweet on the StockNet-dataset for comparison \cite{bertweet}. For the inputs in this experiment, BERTweet can only process the immediate Tweets before the target day, $X_{t-1}$. The StockNet model can process the textual information and price information over the lag periods, those being $X$ and $M$. TEANet $X$ and $M$ in their entirety as well, putting TEANet, StockNet, and MEANT-Tweet on relatively equal footing in terms of their processing capabilities. Experimental settings for each model can be seen in \ref{tab:training_settings}.

StockNet is a binary classification problem, like TempStock. StockNet is built upon price movement. Built over a five day lag period, the classification of labels focused on the price change between the adjusted closing price of the last auxiliary day $d-1$ and the target day $d$, denoted $p_d^c$ and $p_{d-1}^c$ respectively in the original paper \cite{stocknet}. The labels are determined as follows: 

\begin{equation}
y = \mathbbm{1} \left( p_d^c > p_{d-1}^c \right)
\end{equation}

Lag periods that had a movement ratio $r$ where $ -0.5\%  < r \leq 0.55\%$ were thrown out. The movement ratio is calculated as follows:  
\begin{equation}
    r = (p_d^c - p_{d-1}^c)/p_{d-1}^c
\end{equation}
\begin{table}[h]
\centering
\small
\begin{tabular}{@{}lcccc@{}}
\toprule
\textbf{Model}        & \textbf{Task}      & \textbf{epochs} & \textbf{Batch} & \textbf{Patience}\\
\midrule
MEANT-B & TempStock & 15 & 16 & 3\\
M-Tweet-P-base           & StockNet & 10 & 32 & 3 \\
M-Large & TempStock & 15 & 16 & 3 \\
 & MOSI & 15 & 16 & 3 \\
M-Tweet-P-Large & StockNet & 10 & 32& 3\\
M-XL & TempStock &  15 & 16 & 3\\
M-Tweet-P-XL  & StockNet & 10 & 32 & 3 \\
FinBERT & TempStock & 11 & 16 &3 \\
            & StockNet & 7 & 32 &3 \\
BERT & TempStock & 15  & 16 & 3 \\
            & StockNet & 10 & 32  & 3 \\
BERTweet & StockNet & 15 & 16 & 3\\
VL-BERT & TempStock & 15 &  16 & 3\\
ViLT & TempStock & 15 & 16  & 3 \\
TimeSFormer & TempStock & 15 & 16 & 3\\
MLP & TempStock & 15 & 16 & 3\\
    & StockNet & 7 & 32 & 3\\
LSTM & TempStock & 15 & 16 & 3\\
    & StockNet & 4 & 32 & 3\\
TEANet & TempStock & 15 & 32 & 5\\
    & StockNet & 10 & 16 & 3\\

\bottomrule
\end{tabular}
\caption{Training Settings. M refers to MEANT, and P to Price.}
\label{tab:training_settings}
\end{table}

\subsection{Model Implementation Details}
\label{sec:Implementations}
All models were implemented in Pytorch \cite{torch}. MEANT was implemented using a typical transformer formula, employing the use of RMSNorm \cite{rmsnorm}, Flash-attention \cite{flashattention}, and GELU activation units \cite{gelu}. For our TimeSFormer implementation, we decided to use Phil Wangs \cite{lucidrainsrepo}, for its simplicity, readability, and its use of the Einops library \cite{einops}, which we used in our native MEANT implementations.

There is no public implementation available for TEANet \cite{ZHANG2022117239}, so we implemented the model from the details given in the paper. We used the built in torch LSTM implementation, and the FinBERT embedding layers \cite{araci2019finbert} in order to balance against our implementation of MEANT, and to take advantage of the FinBERT tokenizer. 

For all of our BERT-based encoder models, we used the implementations from the transformer models \cite{hugging_face}. 

\begin{table}[h]
\centering
\small
\begin{tabular}{@{}lc@{}}
\toprule
\textbf{Model}         & \textbf{Parameter Count}\\
\midrule
MLP & 3,400,642 \\ 
LSTM & 16,400,642 \\
VL-BERT   & 111,450,624 \\
ViLT   & 111,595,008 \\
BERT & 134,899,968\\
MEANT-base  & 48,304,272 \\
MEANT-large  & 152,367,264 \\
MEANT-XL & 265,890,528\\
\bottomrule
\end{tabular}
\caption{Parameter Counts}
\label{tab:Model parameter Counts}
\end{table}

\subsection{CMU-MOSI}

We also decided to test our model on the CMU Multimodal Opinion-level Sentiment Intensity (MOSI) dataset \cite{mosi}. 

This dataset includes audio, text, and video modalities compiled in 299 annotated video segments collected from YouTube monologue movie reviews. The data forms a binary sentiment analysis classification task. 

For our purposes, we focus on the text and video modalities. We run MEANT on these inputs. 

CMU-MOSI is of interest because it examines videos with aligned text over time. Our vision backbone, the TimeSFormer model, is built for video inputs \cite{timesformer}. We measured MEANT against previous SOTA baselines. TEASEL is a multimodal model that uses a pre-trained RoBerta as a backbone \cite{teasel}, using a CNN to break down the audio signals before coupling those with the text. UniMSE is an encoder-decoder model which breaks down the audio, visual, and textual modalities in fusion layers \cite{unimse}. UniMSE also uses a CNN to process the visual features. MMML is the current SOTA for the CMU-MOSI benchmark. MMML uses cross-modal attention, which is integrated into a fusion network \cite{mmml}. Interestingly, MMML does not take in visual inputs. The MEANT-large runs below were collected after 15 epochs of training, using the same optimizer and lr scheduler settings listed above. The other results were taken from previous work \cite{mmml}.

\begin{table}[h]
\centering
\small
\begin{tabular}{@{}lccc@{}}
\toprule
\textbf{CMU-MOSI Results}      & \textbf{$F1_{non0}$}     &  \textbf{$F1_{has0}$}  & \textbf{$ACC_{2has0}$} \\
\midrule
TEASEL  & 85 & 84.72 & 84.79 \\
UniMSE  & 86.42 & 85.83 & 85.85 \\
MMML & 89.67  & 87.45 & 87.51 \\
\addlinespace
MEANT-large & 71.43  & 70.30 & 70.32 \\
\bottomrule
\end{tabular}
\caption{Mosi-dataset experiment results using Precision (P), Recall (R), F-1 scores and testing accuracy (Acc).}
\label{tab:mosi_results}
\end{table}

Looking at the results above, MEANT-large performs considerably worse then previous SOTA benchmarks on the MOSI task. The disparity is expected. \textit{Query-Targeting} in MEANT is designed to put great emphasis on the final component in the information period. In the CMU-MOSI task, this refers to the final frame in the video clip, along with the final text token, which have been aligned. The clips in the dataset are short movie reviews. The final frame in these clips does not contain significant information as to the entire clip \cite{mosi}, in the manner that the final price day in a lag period does to a stock price \cite{ZHANG2022117239} \cite{stocknet}. 

Furthermore, the previous state of the art benchmarks are designed to handle the audio component, which is better aligned to the textual inputs then the video embeddings \cite{mosi}. MEANT was working off of the visual and textual inputs alone. Thus, the performance we do achieve speaks to the soundness of our current architecture. 

We did run TEASEL and UniMSE on TempStock, replacing the audio inputs to their CNNs with our graphical data $G$. We changed the models to support our price data $M$. They were trained over 15 epochs, and train batch size of 16, and all other experimental settings identical to those used in the original TempStock experiments.

\begin{table}[h]
\centering
\small
\begin{tabular}{@{}lccc@{}}
\toprule
\textbf{MOSI models on TempStock}      & \textbf{$F1$}     &  \textbf{$P$}  & \textbf{$R$} \\
\midrule
TEASEL  & 0.6228 & 0.6148 & 0.5745 \\
UniMSE  & 0.7343 & 0.7238 & 0.7315 \\
\addlinespace
MEANT-large & 0.8351 & 0.8399 & 0.8343 \\
\bottomrule
\end{tabular}
\caption{Models which performed well on MOSI, ran on TempStock. Results use Precision (P), Recall (R), F-1 scores and testing accuracy (Acc).}
\label{tab:temp_mosi_results}
\end{table}

 The models which perform at such a high level on MOSI fail to perform as well on the TempStock task, as seen in \ref{tab:temp_mosi_results}. Ideally, one architecture could tackle both of these sorts of problems. In future work, we would like to make our temporal mechanism more robust to dependencies across the time dimension of the entire input. One method would be to extend our \textit{Query-Targeting} mechanism to learn a parameterized selection of the best target components, or to learn which parts of the input the other auxiliary dependencies need to be collected in relation to. This could involve a separate temporal matrix, as in \citet{tempLLM}, or some sort of softmax query weighting prior to the attention computation. Creating a mechanism which can perform at the highest level on any temporally dependent benchmark remains an open problem. 

\subsection{TempStock Dataset Details}
\label{sec:TempStock_data_details}
The tables below show the number of lag periods used in TempStock for each ticker.

\begin{table}[t]
\centering
\begin{tabular}{@{}lc@{}}
\toprule
\textbf{Ticker} & \textbf{Count} \\
\midrule
DHI & 32 \\
HWM & 31 \\
PCG & 30 \\
LEN & 28 \\
DG & 28 \\
IR & 27 \\
EL & 27 \\
AVGO & 26 \\
CTRA & 26 \\
IEX & 26 \\
XRAY & 26 \\
TER & 26 \\
KR & 26 \\
UPS & 25 \\
PAYC & 25 \\
META & 25 \\
L & 25 \\
PGR & 25 \\
FITB & 25 \\
BKR & 25 \\
LYV & 25 \\
DRI & 25 \\
MET & 25 \\
WYNN & 25 \\
SHW & 25 \\
APTV & 25 \\
SEE & 25 \\
AMCR & 24 \\
ADI & 24 \\
ANSS & 24 \\
HUM & 24 \\
DXC & 24 \\
CRM & 24 \\
SBNY & 24 \\
STLD & 24 \\
CMI & 24 \\
PWR & 24 \\
MKTX & 24 \\
LUV & 24 \\
REGN & 24 \\
RTX & 24 \\
MNST & 24 \\
CDW & 24 \\
MHK & 24 \\
VRTX & 24 \\
TMUS & 23 \\
TRGP & 23 \\
WAB & 23 \\
APH & 23 \\
FTNT & 23 \\
GRMN & 23 \\
FDX & 23 \\
\bottomrule
\end{tabular}
\caption{TempStock Companies Chunk 1}
\label{tab:TempStock_results}
\end{table}
\begin{table}[t]
\centering
\begin{tabular}{@{}lc@{}}
\toprule
\textbf{Ticker} & \textbf{Count} \\
\midrule
FE & 23 \\
JNPR & 23 \\
INTU & 23 \\
HBAN & 23 \\
NOC & 23 \\
CLX & 23 \\
LVS & 23 \\
SBUX & 23 \\
JPM & 23 \\
NOW & 23 \\
DGX & 23 \\
LOW & 23 \\
PNC & 23 \\
PPG & 23 \\
ECL & 23 \\
ZTS & 23 \\
TMO & 23 \\
XYL & 23 \\
EPAM & 22 \\
DAL & 22 \\
LUMN & 22 \\
MRO & 22 \\
MGM & 22 \\
MTCH & 22 \\
ENPH & 22 \\
HSY & 22 \\
GIS & 22 \\
OTIS & 22 \\
NRG & 22 \\
WRB & 22 \\
EVRG & 22 \\
NDSN & 22 \\
NVR & 22 \\
CHD & 22 \\
CBOE & 22 \\
HCA & 22 \\
CDNS & 22 \\
SWKS & 22 \\
PEP & 22 \\
LW & 22 \\
TYL & 21 \\
RL & 21 \\
SWK & 21 \\
FANG & 21 \\
PTC & 21 \\
QCOM & 21 \\
DUK & 21 \\
MTD & 21 \\
AEP & 21 \\
LLY & 21 \\
MMM & 21 \\
ABT & 21 \\
\bottomrule
\end{tabular}
\caption{TempStock Companies Chunk 2}
\label{tab:TempStock_results}
\end{table}

\begin{table}[t]
\centering
\begin{tabular}{@{}lc@{}}
\toprule
\textbf{Ticker} & \textbf{Count} \\
\midrule
ZBH & 21 \\
UNP & 21 \\
TSCO & 21 \\
TFC & 21 \\
LHX & 21 \\
HIG & 21 \\
HON & 21 \\
KEYS & 21 \\
KDP & 21 \\
CBRE & 21 \\
CMS & 21 \\
MSFT & 21 \\
NSC & 21 \\
VMC & 21 \\
AIG & 21 \\
GM & 21 \\
FOX & 21 \\
BAC & 21 \\
TTWO & 21 \\
BIO & 21 \\
ETSY & 21 \\
ZION & 20 \\
MCK & 20 \\
NVDA & 20 \\
CHRW & 20 \\
CAG & 20 \\
LKQ & 20 \\
BBY & 20 \\
BIIB & 20 \\
HLT & 20 \\
NEM & 20 \\
CCI & 20 \\
FTV & 20 \\
CARR & 20 \\
ODFL & 20 \\
PCAR & 20 \\
WBA & 20 \\
PEG & 20 \\
PSX & 20 \\
HII & 20 \\
GL & 20 \\
SJM & 20 \\
CI & 20 \\
FSLR & 20 \\
TJX & 20 \\
MAR & 20 \\
CSGP & 20 \\
UAL & 20 \\
T & 20 \\
SNPS & 20 \\
AEE & 20 \\
DTE & 20 \\
\bottomrule
\end{tabular}
\caption{TempStock Companies Chunk 3}
\label{tab:TempStock_results}
\end{table}

\begin{table}[t]
\centering
\begin{tabular}{@{}lc@{}}
\toprule
\textbf{Ticker} & \textbf{Count} \\
\midrule
ETN & 20 \\
WHR & 20 \\
GOOGL & 19 \\
GOOG & 19 \\
SYK & 19 \\
DLR & 19 \\
AES & 19 \\
ADP & 19 \\
AIZ & 19 \\
ADSK & 19 \\
AKAM & 19 \\
KEY & 19 \\
TRMB & 19 \\
UDR & 19 \\
JNJ & 19 \\
IBM & 19 \\
ILMN & 19 \\
CF & 19 \\
SCHW & 19 \\
CB & 19 \\
CINF & 19 \\
PAYX & 19 \\
PYPL & 19 \\
IVZ & 19 \\
FOXA & 19 \\
EFX & 19 \\
OXY & 19 \\
TECH & 19 \\
VRSK & 19 \\
HPE & 19 \\
NDAQ & 19 \\
NTRS & 19 \\
CNC & 19 \\
CMA & 19 \\
CSCO & 19 \\
ALL & 19 \\
ABBV & 19 \\
LNT & 19 \\
VFC & 18 \\
VTRS & 18 \\
AAL & 18 \\
AMGN & 18 \\
YUM & 18 \\
CEG & 18 \\
C & 18 \\
ON & 18 \\
NKE & 18 \\
NXPI & 18 \\
AAP & 18 \\
EXR & 18 \\
EQT & 18 \\
CE & 18 \\
\bottomrule
\end{tabular}
\caption{TempStock Companies Chunk 4}
\label{tab:TempStock_results}
\end{table}

\begin{table}[t]
\centering
\begin{tabular}{@{}lc@{}}
\toprule
\textbf{Ticker} & \textbf{Count} \\
\midrule
ORLY & 18 \\
JCI & 18 \\
MPC & 18 \\
CVS & 18 \\
GE & 18 \\
K & 18 \\
TXN & 18 \\
HD & 18 \\
MOS & 18 \\
CVX & 18 \\
CL & 18 \\
HPQ & 18 \\
ITW & 18 \\
WMT & 18 \\
PM & 18 \\
MU & 18 \\
MPWR & 18 \\
MSCI & 18 \\
MAS & 18 \\
TEL & 18 \\
BAX & 18 \\
VZ & 18 \\
WMB & 18 \\
SLB & 18 \\
DFS & 18 \\
WST & 18 \\
MCD & 18 \\
MRK & 18 \\
DXCM & 18 \\
SYY & 18 \\
AMAT & 18 \\
AFL & 17 \\
A & 17 \\
MRNA & 17 \\
NTAP & 17 \\
NWSA & 17 \\
NEE & 17 \\
MAA & 17 \\
CSX & 17 \\
DHR & 17 \\
IRM & 17 \\
J & 17 \\
DE & 17 \\
CPT & 17 \\
OGN & 17 \\
ED & 17 \\
LIN & 17 \\
CAT & 17 \\
BSX & 17 \\
F & 17 \\
BEN & 17 \\
EXPD & 17 \\
\bottomrule
\end{tabular}
\caption{TempStock Companies Chunk 5}
\label{tab:TempStock_results}
\end{table}

\begin{table}[t]
\centering
\begin{tabular}{@{}lc@{}}
\toprule
\textbf{Ticker} & \textbf{Count} \\
\midrule
EA & 17 \\
EOG & 17 \\
CTSH & 17 \\
KLAC & 17 \\
CMG & 17 \\
FCX & 17 \\
FMC & 17 \\
IPG & 17 \\
BK & 17 \\
BKNG & 17 \\
TROW & 17 \\
PNR & 17 \\
CRL & 17 \\
WAT & 17 \\
WFC & 17 \\
RMD & 17 \\
BLK & 17 \\
EIX & 17 \\
EW & 17 \\
D & 17 \\
WDC & 17 \\
STX & 17 \\
SNA & 17 \\
RHI & 17 \\
SBAC & 17 \\
V & 17 \\
AXP & 17 \\
AMT & 17 \\
VLO & 16 \\
PSA & 16 \\
BBWI & 16 \\
BDX & 16 \\
TGT & 16 \\
TDY & 16 \\
WRK & 16 \\
WY & 16 \\
WTW & 16 \\
XEL & 16 \\
WBD & 16 \\
TSN & 16 \\
LRCX & 16 \\
LMT & 16 \\
BMY & 16 \\
GPN & 16 \\
GS & 16 \\
HSIC & 16 \\
CTVA & 16 \\
LYB & 16 \\
MA & 16 \\
GPC & 16 \\
GILD & 16 \\
CTLT & 16 \\
\bottomrule
\end{tabular}
\caption{TempStock Companies Chunk 6}
\label{tab:TempStock_results}
\end{table}

\begin{table}[t]
\centering
\begin{tabular}{@{}lc@{}}
\toprule
\textbf{Ticker} & \textbf{Count} \\
\midrule
ROK & 16 \\
MKC & 16 \\
ADM & 16 \\
ACGL & 16 \\
ANET & 16 \\
AZO & 16 \\
ALLE & 16 \\
ELV & 16 \\
ETR & 16 \\
EXC & 16 \\
XOM & 16 \\
EMR & 15 \\
EQR & 15 \\
ESS & 15 \\
ZBRA & 15 \\
ACN & 15 \\
ATO & 15 \\
AMP & 15 \\
CTAS & 15 \\
PARA & 15 \\
ROP & 15 \\
GLW & 15 \\
MTB & 15 \\
MLM & 15 \\
DPZ & 15 \\
DD & 15 \\
NFLX & 15 \\
BRO & 15 \\
KMX & 15 \\
GD & 15 \\
USB & 15 \\
SRE & 15 \\
STT & 15 \\
CME & 15 \\
CMCSA & 15 \\
INCY & 15 \\
IFF & 15 \\
RSG & 15 \\
FDS & 15 \\
BWA & 15 \\
BXP & 15 \\
TFX & 15 \\
NI & 15 \\
NUE & 15 \\
ORCL & 15 \\
PNW & 15 \\
PLD & 15 \\
IT & 15 \\
AVB & 15 \\
AWK & 15 \\
AJG & 14 \\
\bottomrule
\end{tabular}
\caption{TempStock Companies Chunk 7}
\label{tab:TempStock_results}
\end{table}

\begin{table}[t]
\centering
\begin{tabular}{@{}lc@{}}
\toprule
\textbf{Ticker} & \textbf{Count} \\
\midrule
UHS & 14 \\
VICI & 14 \\
FIS & 14 \\
GEN & 14 \\
NCLH & 14 \\
DLTR & 14 \\
IP & 14 \\
INVH & 14 \\
RJF & 14 \\
NWL & 14 \\
HST & 14 \\
PKG & 14 \\
CPB & 14 \\
COF & 14 \\
GNRC & 14 \\
EMN & 14 \\
MCHP & 14 \\
MDLZ & 14 \\
MS & 14 \\
CNP & 14 \\
PH & 14 \\
CCL & 14 \\
DVA & 14 \\
DVN & 14 \\
SPG & 14 \\
TSLA & 14 \\
ULTA & 14 \\
KMB & 14 \\
KHC & 14 \\
RCL & 14 \\
BALL & 14 \\
SYF & 14 \\
APD & 13 \\
MO & 13 \\
AMZN & 13 \\
AVY & 13 \\
EQIX & 13 \\
CZR & 13 \\
JBHT & 13 \\
PXD & 13 \\
VNO & 13 \\
RF & 13 \\
PFE & 13 \\
ISRG & 13 \\
ICE & 13 \\
INTC & 13 \\
LDOS & 13 \\
COST & 13 \\
MDT & 13 \\
COP & 13 \\
AON & 13 \\
AOS & 13 \\
\bottomrule
\end{tabular}
\caption{TempStock Companies Chunk 8}
\label{tab:TempStock_results}
\end{table}

\begin{table}[t]
\centering
\begin{tabular}{@{}lc@{}}
\toprule
\textbf{Ticker} & \textbf{Count} \\
\midrule
STE & 13 \\
VTR & 13 \\
WM & 13 \\
DIS & 13 \\
JKHY & 12 \\
PG & 12 \\
IQV & 12 \\
AMD & 12 \\
FRT & 12 \\
ALB & 12 \\
KIM & 12 \\
MSI & 12 \\
ROST & 12 \\
URI & 12 \\
ES & 12 \\
HRL & 12 \\
O & 12 \\
GWW & 12 \\
PFG & 12 \\
PPL & 12 \\
ADBE & 12 \\
HAL & 12 \\
DOW & 12 \\
ARE & 11 \\
BR & 11 \\
IDXX & 11 \\
AAPL & 11 \\
BA & 11 \\
OKE & 11 \\
VRSN & 11 \\
WELL & 11 \\
TDG & 11 \\
SPGI & 11 \\
EBAY & 11 \\
CPRT & 11 \\
UNH & 10 \\
SO & 10 \\
NWS & 10 \\
KMI & 10 \\
REG & 9 \\
DOV & 9 \\
HES & 1 \\
\bottomrule
\end{tabular}
\caption{TempStock Companies Chunk 9}
\label{tab:TempStock_results}
\end{table}
\end{document}